\def\year{2018}\relax
\documentclass[letterpaper]{article} 
\usepackage{aaai18}  
\usepackage{times}  
\usepackage{helvet}  
\usepackage{courier}  
\usepackage{url}  
\usepackage{graphicx}  

\usepackage{epsfig}
\usepackage{graphicx}
\usepackage{amsmath}
\usepackage{amssymb}
\usepackage{bm}
\usepackage{algorithmic}
\usepackage{graphicx}
\usepackage{subfigure}
\usepackage{multirow}%
\usepackage{amsmath, amssymb}
\usepackage{color}
\usepackage{mathrsfs}
\usepackage{amsthm}
\usepackage{epstopdf}
\usepackage{wrapfig}
\usepackage{picinpar}
\usepackage{url}
\usepackage{algorithm}

\graphicspath{{figures/}}

\def\mB{{\mathcal B}}

\def\0{{\bf 0}}
\def\1{{\bf 1}}

\def\bx{{\bf x}}
\def\by{{\bf y}}

\def\citep{\cite}
\def\citet{\cite}

\begin{document}

\title{Double Forward Propagation for Memorized Batch Normalization}
\author{
Yong Guo\thanks{Authors contributed equally.}, Qingyao Wu$^*$, Chaorui Deng, Jian Chen, and Mingkui Tan\thanks{Corresponding author.} \\
School of Software Engineering, South China University of Technology, China \\
\{guo.yong, secrdyz\}@mail.scut.edu.cn, \{qyw, ellachen, mingkuitan\}@scut.edu.cn
}

\maketitle
\begin{abstract}
Batch Normalization (BN) has been a standard component in designing deep neural networks (DNNs). Although the standard BN can significantly accelerate the training of DNNs and improve the generalization performance, it has several underlying limitations which may hamper the performance in both training and inference.
In the training stage, BN relies on estimating the mean and variance of data using a single mini-batch.
Consequently, BN can be unstable when the batch size is very small or the data is poorly sampled.
In the inference stage, BN often uses the so called moving mean and moving variance instead of batch statistics, i.e., the training and inference rules in BN are not consistent.
Regarding these issues, we propose a memorized batch normalization (MBN), which considers multiple recent batches to obtain more accurate and robust statistics.
Note that after the SGD update for each batch, the model parameters will change, and the features will change accordingly, leading to the \emph{Distribution Shift} before and after the update for the considered batch.
To alleviate this issue, we present a simple Double-Forward scheme in MBN which can further improve the performance. Compared to related methods, the proposed MBN exhibits consistent behaviors in both training and inference. Empirical results show that the MBN based models trained with the Double-Forward scheme greatly reduce the sensitivity of data and significantly improve the generalization performance.
\end{abstract}

\section{Introduction}\label{sec:introduction}
Deep neural networks (DNNs) have become the workhorse of many learning tasks and real-world applications including computer vision~\cite{krizhevsky2012imagenet,he2016deep,szegedy2017inception}, natural language understanding~\cite{goldberg2016primer,vendrov2015order,zhou2017natural} and speech recognition~\cite{lecun2015deep,xiong2017microsoft,heymann2016neural}. Training very deep DNNs is an important problem and comes with its own set of challenges. One of the most challenging issues is brought by the so called \emph{Internal Covariate Shift}~\cite{ioffe2015batch}, which refers to the data distribution changes for each layer during training. Specifically, when training DNNs with mini-batch stochastic gradient descent (SGD)~\cite{bottou1998online}, the input of each layer will change due to the update of model parameters of the previous layers. As a result, the gradient vanishing or exploding problems may occur with a large probability, which severely hampers the performance~\cite{glorot2010understanding,srivastava2015highway}.

To address this issue, Ioffe and Szegedy~\cite{ioffe2015batch} proposed the Batch Normalization (BN) method, which seeks to reduce the \emph{Internal Covariate Shift} by applying data normalization for the input of each layer. The data in each batch $\mB$ can be normalized by
\begin{equation}\label{eqn:batchnorm}
\hat \bx_i = \frac{\bx_i - \mu_\mB} {\sigma_\mB},~~\by_i = \gamma \hat \bx_i + \beta,
\end{equation}
where $\mu_\mB$ and $\sigma_\mB$ are the mean and variance of the data batch, and $\gamma$ and $\beta$ are learnable parameters to restore the representative power of the model.
In fact, by setting $\gamma$ to $\sigma_\mB$ and $\beta$ to $\mu_\mB$, one can recover the original input $\bx_i$.
When doing the inference, however, the statistics of a single mini-batch can be very unstable due to insufficient data or the unreliable data quality. To address this, one may average the statistics of more batches. For example, in many deep learning packages, such as {PyTorch and TensorFlow,} the so called moving mean $\mu_{\mathrm{mov}}$ and moving variance $\sigma_{\mathrm{mov}}$ are used for inference. By moving averages over all the training batches, $\mu_{\mathrm{mov}}$ and $\sigma_{\mathrm{mov}}$ can be computed recursively via
\small
\begin{equation}\label{moving_aveage}
\mu_{\mathrm{mov}} := \theta \mu_{\mathcal B} + (1-\theta)\mu_{\mathrm{mov}},~~~
\sigma_{\mathrm{mov}} := \theta \sigma_{\mathcal B} + (1-\theta)\sigma_{\mathrm{mov},}
\end{equation}
\normalsize
where $\theta\leq 1$ is the coefficient of the linear combination.

Batch normalization can significantly accelerate the training of DNNs by effectively addressing the gradient vanishing issue.
Moreover, as BN is able to reduce the distribution bias of each layer, we can train deep models with a large learning rate and without the need of careful initialization~\cite{he2016deep}. Batch normalization has also been extended to other models such as recurrent neural networks(RNNs)~\cite{laurent2016batch,amodei2016deep,cooijmans2016recurrent} and successfully been applied on many challenging tasks~\cite{ba2016layer,dumoulin2017learned,li2016revisiting,tang2017train,zhang2017deep}.

However, the standard BN which relies on a single batch of data has some underlying limitations.
\textbf{First}, the training and inference are inconsistent, which means that the learned rules are not optimal.
\textbf{Second}, the mean and variance computed on a small batch can be inaccurate and unstable. In this sense, batch normalization with a single batch may fail when the batch size is very small (even a single sample)~\cite{ba2016layer,salimans2016weight}.
\textbf{Moreover}, the normalization is sensitive to data quality. Even if the data batch is sufficiently large, the input distribution can still vary greatly due to the strong dependency on sampled data at each iteration.

To amend the above issues, a straightforward approach is to take more batches to estimate the data statistics, such as the moving mean $\mu_{\mathrm{mov}}$ and moving variance $\sigma_{\mathrm{mov}}^2$~\cite{dinh2016density}.
However, directly applying moving mean and moving variance may incur convergence issues and blow up the model~\cite{ioffe2015batch}.
One possible reason, as argued in \cite{ioffe2015batch,ioffe2017batch}, is that while gradient step is designed to decrease the loss, the normalization step would cancel
the effect of these changes in the loss, leading to little performance improvement but an unbounded growth of model parameters.
Another reason can be the inaccurate moving mean and variance, which are simple weighted sum of means and variances at previous iterations.
Instead, we propose a memorized batch normalization (MBN) method which directly considers the mean and variance of the data from multiple batches.

In MBN, we consider multiple batches, but pay particular attention on the current batch and the next batch. In this setting,
the current batch will have non-negligible distribution shift
when going to the next iteration due to the update of model parameters,
which will lead to inaccurate estimation of mean and variance for the next iteration.
Regarding this issue, we present a Double-Forward training scheme by simply performing one additional forward propagation, which can further improve the learning performance.

In this paper, we make the following contributions:
\begin{itemize}
\item We propose a memorized batch normalization (MBN) method which considers data information from multiple recent batches (or all batches in an extreme case) rather than a single batch to produce more accurate and stable statistics. Since MBN considers data information from multiple batches, SGD with a very small batch sizes can be still applied to train DNNs.
\item  We propose a Double-Forward scheme to address the issue of distribution shift among iterations in MBN. Thus, the estimation of mean and variance at the current iteration is
    kept up-to-date, which benefits the training at the next iteration.
\item Based on the proposed MBN, the training and inference of DNNs will share the same normalization and forward propagation rules.  Empirical results show that, equipped with the proposed MBN and training scheme, DNN models exhibit improved generalization performance compared to several state-of-the-arts.
\end{itemize}

\section{Related Studies}
To alleviate \emph{Internal Covariate Shift}, one can whiten the output of each layer, but it turns out to be very computationally expensive~\cite{wiesler2014mean,raiko2012deep}. Batch Normalization (BN) \cite{ioffe2015batch} approximates the whitening by normalizing the layer activations using data statistics of the current mini-batch. However, while BN only considers single batch statistics during training, the moving mean and moving variance are utilized for inference, which brings in a representation gap between training and inference.

Another kind of methods conducts normalization using the model parameters rather than data statistics, such as Weight Normalization (WN)~\cite{salimans2016weight} and Normalization Propagation (NormProp)~\cite{arpit2016normalization}. WN proposes to use model parameters to normalize the layer activations but does not guarantee its effectiveness on nonlinear components.
NormProp further extends the parameter-based normalization method to more kinds of operations, including convolution, linear transformation, and ReLU activation~\cite{nair2010rectified}, by performing layer-wise normalization.
However, the assumption that the input of each layer has to follow $N(0,1)$ is too strong. Especially for ReLU, after discarding the negative part of samples, the residual positive part does not follow the normal distribution. Thus, the input of the next layer will not satisfy this assumption. Moreover, the harmful effects of failing the assumption will accumulate as the model goes deeper.

Recently, performing normalization using multiple batches has been studied in several works.
AdaBN~\cite{li2016revisiting} apply data normalization over all the samples using a pre-trained model for domain adaptation.
~\cite{dinh2016density} proposes to combine the current batch statistics with the moving averages to stabilize the training with data normalization.
While preparing this work, we were aware that Ioffe had proposed Batch Renormalization (BRN)~\cite{ioffe2017batch} method very recently, which has similar motivations with ours.
Batch Renormalization aims to gradually make use of popular statistics $\mu_{\mathrm{mov}}, \sigma_{\mathrm{mov}}$ by projecting the single batch normalized values $\hat \bx_i$ to its moving averages normalized values $\by_i$ during training, which can be computed by an affine function:
\begin{equation}\label{eq:affine}
\by_i = r\hat \bx_i + d,
\end{equation}
with $r$ and $d$ being set to $\it{clip}_{[1/ {r_{\mathrm{max}}}, r_{\mathrm{max}}]} \left(\frac{\sigma_{\mathcal B}} {\sigma_{\mathrm{mov}}} \right)$ and $\it{clip}_{[-d_{\mathrm{max}}, d_{\mathrm{max}}]}\frac{\mu_{\mathcal B} - \mu_{\mathrm{mov}}} {\sigma_{\mathrm{mov}}}$, respectively.
Specifically, the parameters are set to $r_{\mathrm{max}}=1, d_{\mathrm{max}}=0$ at the early training stage, which makes the affine function an identity mapping, and reduces the BRN to the original Batch Normalization. Then the parameters are gradually relaxed to $r_{\mathrm{max}}=3, d_{\mathrm{max}}=5$. However, BRN still uses the moving averages during inference. In this sense, once the clip operations take effect, the training and inference will show inconsistent behaviors.
Moreover, Batch Renormalization uses a ``fairly high" decay rate $\theta$ in Eqn.~({\ref{moving_aveage}}) to update the moving averages, thus these moving averages can be very close to the current batch statistics $\mu_{\mathcal B}, \sigma_{\mathcal B}$ and BRN can be close to BN.
Unlike BRN, our method takes multiple batches to perform normalization during the whole training procedure and keep the consistent behaviors in forward and backward propagations.

\section{Memorized Batch Normalization}\label{sec:MBN}
In this section, we propose to use memorized statistics over multiple batches for data normalization. After that, to alleviate the distribution shift issue incurred by SGD update, we present an effective Double-Forward training scheme.

\subsection{Memorized Statistics}
Normalization depending on single data batch can be inaccurate and unstable when the batch size is very small or the data is poorly sampled~\cite{li2016revisiting}. A natural solution  is to take multiple batches into account to obtain more accurate mean and variance~\cite{dinh2016density}. Specifically,~\cite{dinh2016density} performs the normalization by combining the batch mean and batch variance with the moving mean and moving variance:
\begin{equation}\label{eq:memorized-statistics}
{{\bf{\mu_{\mathrm{mov}}}}} = \frac{\sum_{i=1}^{T} {\alpha_i \cdot n_i \mu_i}}{\sum_{i=1}^{T} {\alpha_i \cdot n_i}},~~
 \sigma_{\mathrm{mov}}^2 = \frac{\sum_{i=1}^{T} \alpha_i \cdot n_i  \sigma_i^2 } {\sum_{i=1}^{T} {\alpha_i \cdot n_i}},
\end{equation}
where $T$ is the number of iterations, $\alpha_i$ is the weight for the $i$-th batch and $n_i$ denotes the number of samples in the batch.
However, performing normalization directly using moving mean and moving variance may incur convergence issue~\cite{ioffe2015batch}.

Differently, we propose a memorized batch normalization (MBN) which directly considers the mean and variance of the data from multiple batches.
Specifically, MBN considers the data of the current batch and the recent $k$ batches, which means that we take $(k+1)$ batches as a large batch to compute the corresponding mean and variance. Then, these estimations are utilized to normalize the layer activations for the next iteration.
Let $\bx_i^{(j)}$ be the $j$-th sample of the $i$-th batch.
Then, the memorized statistics for the data of all $(k+1)$ batches can be computed by
\begin{small}
\begin{equation}\label{eq:memorized-statistics1}
{\hat{\bf{\mu}}} =
\frac{\sum_{i = 1}^{k+1} \alpha_i{\sum_{j=1}^{n_i} {\bx_i^{(j)}}}}{\sum_{i=1}^{k+1}\alpha_i {n_i}},
\hat \sigma^2 = \frac{\sum_{i=1}^{k+1} {\sum_{j=1}^{n_i} {\alpha_i (\bx_i^{(j)}-\hat \mu)^2}}} {\sum_{i=1}^{k+1}\alpha_i {n_i}},
\end{equation}
\end{small}
where $\hat \mu$ and $\hat \sigma^2$ denote the mean and variance, respectively. Here, $\alpha_i$
denotes the weight of the $i$-th pair of mean and variance in memory.
And $\alpha_{k+1}=1$ is set for the current batch mean and batch variance by default.
Eqn.~(\ref{eq:memorized-statistics1}) can be equivalently transformed into the following form:
\begin{small}
\begin{equation}\label{eq:memorized-statistics2}
{\hat{\bf{\mu}}} = \frac{\sum_{i=1}^{k+1} {\alpha_i \cdot n_i \mu_i}}{\sum_{i=1}^{k+1} {\alpha_i \cdot n_i}},~~
\hat \sigma^2 = \frac{\sum_{i=1}^{k+1} \alpha_i \cdot n_i \left( \left( \mu_i-\hat \mu \right)^2 + \sigma_i^2 \right)} {\sum_{i=1}^{k+1} {\alpha_i \cdot n_i}},
\end{equation}
\end{small}
where $\mu_i$ and $\sigma_i$ denotes the estimations of mean and variance for the $i$-th batch.
Note that if the batch size is kept the same for all the batches, the $n_i$ in (\ref{eq:memorized-statistics2}) can be cancelled out.
Different from (\ref{eq:memorized-statistics}),  the variance $\hat \sigma^2$ in (\ref{eq:memorized-statistics2}) considers the statistics among batches with a \emph{correction} term $\left( \mu_i-\hat \mu \right)^2$. The correction term considers the distribution shift of previous iterations, which will alleviate the estimation bias of mean and variance for the current SGD update.

According to Eqn.~(\ref{eq:memorized-statistics2}), when computing $\hat{\bf{\mu}}$ and $\hat \sigma^2$,  we only need to record the mean and variance in memory rather than the whole feature map of each layer. Therefore, the computation of $\hat \mu$ and $\hat \sigma$ is quite simple and MBN exhibits comparable time complexity to BN.

Similar to BN, in MBN, we also use a learnable pair of parameters $\gamma$ and $\beta$ to conduct normalization by
\begin{equation}\label{eq:inference}
\hat \bx_i^{(j)} = \frac{\bx_i - \hat \mu} {\hat \sigma}, ~~\by_i^{(j)} = \gamma \hat \bx_i^{(j)} + \beta.
\end{equation}
The parameters $\gamma$ and $\beta$ are capable of restoring the original representations if the model is not optimal.

\textbf{Backward Propagation}. When doing the training, we need to propagate the gradients through MBN transformation via backward propagation. Based on the chain rule, we can compute the gradients of MBN for the data of the current batch $\mathcal B_{k+1}$ as follows:

\begin{small}
\begin{equation}
\left.\begin{aligned}
&\frac{\partial \ell} {\partial \bx_{k+1}^{(j)}} = \frac{\partial \ell} {\partial \by_{k+1}^{(j)}} \cdot \gamma, \\
&\frac{\partial \ell} {\partial \hat \sigma^2} = - \frac{1}{2} \sum\nolimits_{j=1}^{n_{k+1}} {\frac{\partial \ell} {\partial \bx_{k+1}^{(j)}} (\bx_k^{(j)} - \hat \mu) (\hat \sigma^2 + \epsilon)^{-3/2}}, \\
&\frac{\partial \ell} {\partial \hat \mu} = \sum\nolimits_{j=1}^{n_{k+1}} \frac{\partial \ell} {\partial \bx_{k+1}^{(j)}} \frac{-1} {\sqrt{\hat \sigma^2 + \epsilon}}
- 2 \frac{\partial \ell} {\partial \hat \sigma^2} \frac{\sum \nolimits_{i=1}^{k+1} {\alpha_i n_i (\mu_i - \hat \mu)}} {\sum\nolimits_{i=1}^{k+1} {\alpha_i n_i}} ,\\
&\frac{\partial \ell} {\partial \bx_{k+1}^{(j)}}
= \frac{\partial \ell} {\partial \hat \bx_{k+1}^{(j)}} \frac{1} {\sqrt{\hat \sigma^2 + \epsilon}}
+ \frac{\partial \ell} {\partial \hat \sigma^2} \frac{2} {\sum\nolimits_{i=1}^{k+1} {\alpha_i n_i}} \left( \bx_{k+1}^{(j)} -\hat \mu \right) \\
&+ \frac{\partial \ell} {\partial \hat \mu} \frac{1} {\sum\nolimits_{i=1}^{k+1} {\alpha_i n_i}} ,\\
&\frac{\partial \ell} {\partial \gamma} = \sum\nolimits_{j=1}^{n_{k+1}} {\frac{\partial \ell} {\partial \by_{k+1}^{(j)}} \bx_{k+1}^{(j)}} ,\\
&\frac{\partial \ell} {\partial \beta} = \sum\nolimits_{j=1}^{n_{k+1}} {\frac{\partial \ell} {\partial \by_{k+1}^{(j)}}}.
\end{aligned}\right. \nonumber
\end{equation}
\end{small}

\textbf{Forward Propagation and Inference}. In MBN, once the parameters $\hat{\bf{\mu}}$, $\hat \sigma^2$,  $\gamma$ and $\beta$ are computed, they will be fixed and used for the Forward Propagation in the training stage or in the testing stage as in (\ref{eq:inference}). In other words, in MBN, the training and inference become consistent since they share the same rules.

\subsection{Dynamic Setting of Weights}\label{sec:weightsetting}
Due to the distribution shift incurred by SGD update, the more recent batch should be more important for the current iteration. We thus assign larger weights $\alpha$ on the more recent batches. Similar to~\cite{dinh2016density}, we set $$\alpha_i = {\eta^{k-i}},~~ \forall~1 \leq i\leq k$$ where $\eta \leq 1$ is a weight decaying parameter. Note that when $i=k$, we set $\alpha_k = 1$, which means that the importance of the most recent batch $\mB_{k}$ is equal to the current batch $\mB_{k+1}$. This setting is reasonable in the sense that the distribution shift between adjacent batches is often small.

Another important issue is that, the strength of the \emph{Distribution Shift} among iterations is time dependent. Specifically, the model parameters change more violently at the very beginning of training due to larger step size and/or large magnitude of the gradients. As a result, the statistics of different batches at the earlier stages change more significantly, leading to unstable estimations of the mean and variance. To address this, we introduce an additional weight decaying parameter so that
\begin{eqnarray}\label{eq:decay}
\alpha_i = \lambda \eta^{k-i},
\end{eqnarray}
where $\lambda\leq 1$. When seting $\lambda=0$, MBN is reduced to the standard Batch Normalization.

In practice, we can gradually increase $\lambda$ from a small value after the changing of step size. For example, in the experiments of this paper, we  set the parameter $\lambda=0.1$ at the beginning and then change it to $0.5$ and $0.9$ at 40\% and 60\% of the training procedure. Therefore, we choose $\lambda$ from $\{0.1,0.5,0.9\}$.

\section{Double-Forward Propagation}
Since MBN considers information of multiple data batches, after multiple SGD updates, the distribution shift of batches at previous iterations can accumulate to affect the resultant estimations. Especially when the memory size is very large, the previous recorded means and variances will cause an estimation bias on the mean $\hat \mu$ and variance $\hat \sigma$ for the current iteration.

To address this, we here propose a simple scheme, called the Double-Forward Propagation, as in Algorithm \ref{alg:mbn}. Specifically, we first perform a standard forward-backward propagation to train the model. The estimations of mean and variance before updating the model parameters are used to compute the loss and  gradients of the current SGD update.
After updating the model parameters, we perform a second forward propagation step on the same batch using the updated weights. We then record the updated mean and variance in memory before going to the next iteration. In this way, the distribution shift caused by the change of model parameters can be significantly reduced by keeping the statistics up-to-date.
\begin{algorithm}[t]
\caption{Training MBN in Single Iteration.}
\label{alg:mbn}
\begin{algorithmic}[1]\small
	\REQUIRE  Recorded statistics in memory: \\~~~~~~~~~~~~~~~~~~~~~$\{\mu_i\}_{i=1}^k, \{\sigma_i\}_{i=1}^k$;\\
    ~~~~~~~~~~Mean and variance of the current batch: \\~~~~~~~~~~~~~~~~~~~~~~~~~~~~ $\mu_B, \sigma_B$;\\
    ~~~~~~~~~~Weights for batches in memory: $\{\alpha_i\}_{i=1}^k$;\\
    ~~~~~~~~~~Learnable parameters: $\gamma, \beta$.\\
    We define $\mu_{k+1}=\mu_B, \sigma_{k+1}=\sigma_B$ for convenience.
    \STATE \textbf{First Forward Propagation:}
    \STATE ~~$\hat{\bf{\mu}} \leftarrow \frac{\sum\limits_{i=1}^{k+1} {\alpha_i \cdot n_i \mu_i}}{\sum\limits_{i=1}^{k+1} {\alpha_i \cdot n_i}}$
        ~~~~~~~~~~~~~~~~~~~~~~~~~~\//\// memorized mean \\
    \STATE ~~$\hat \sigma^2 \leftarrow \frac{\sum\limits_{i=1}^{k+1} \alpha_i \cdot n_i \left( \left( \mu_i-\hat \mu \right)^2 + \sigma_i^2 \right)} {\sum\limits_{i=1}^{k+1} {\alpha_i \cdot n_i}}$
        ~\//\// memorized variance \\
    \STATE ~~$\hat \bx_i^{(j)} \leftarrow \frac{\bx_i^{(j)} - \hat \mu} {\sqrt{\hat \sigma^2 + \epsilon}}$
        ~~~~~~~~~\//\// normalization transformation \\
    \STATE ~~$\by_i^{(j)} \leftarrow \gamma \hat \bx_i^{(j)} + \beta$
        ~~~~~~~~~~~~~~~~~~~~~~~~~~~~\//\// scale and shift \\
    \STATE \textbf{Backpropagation:}
    \STATE ~~compute gradients ~~~$\frac{\partial \ell} {\partial \gamma}$, ~~~$\frac{\partial \ell} {\partial \beta}$, ~~~$\frac{\partial \ell} {\partial \bx_{k+1}^{(j)}}$
    \STATE ~~backpropagate $\frac{\partial \ell} {\partial \bx_{k+1}^{(j)}}$ to previous layers
    \STATE ~~$\gamma:=\gamma - \eta \frac{\partial \ell} {\partial \gamma}$, ~~~$\beta:=\beta - \eta \frac{\partial \ell} {\partial \beta}$
    \STATE \textbf{Second Forward Propagation}
    \STATE ~~compute $\mu_B$; record the updated mean of the current batch
    \STATE ~~compute $\sigma_B$; record the updated variance of the current batch
\end{algorithmic}
\end{algorithm}

\section{Computational Complexity}
The computation of memorized statistics in MBN is a simple linear transformation over recorded means and variances, thus MBN has the same cost as BN. In practice, we observe that the backpropagation often takes twice more time than the forward propagation during training. Therefore, in the training,  the Double-Forward scheme with MBN takes about $30\%$ more time than that with BN. For the inference,  MBN takes the same complexity to BN. Nevertheless, MBN based models perform data normalization with more accurate and robust statistical estimations and thus often yield better generalization performance than BN models.

\begin{figure*}[t]
  \centering
      \subfigure[Testing error.]{
      \includegraphics[width = 0.75\columnwidth]{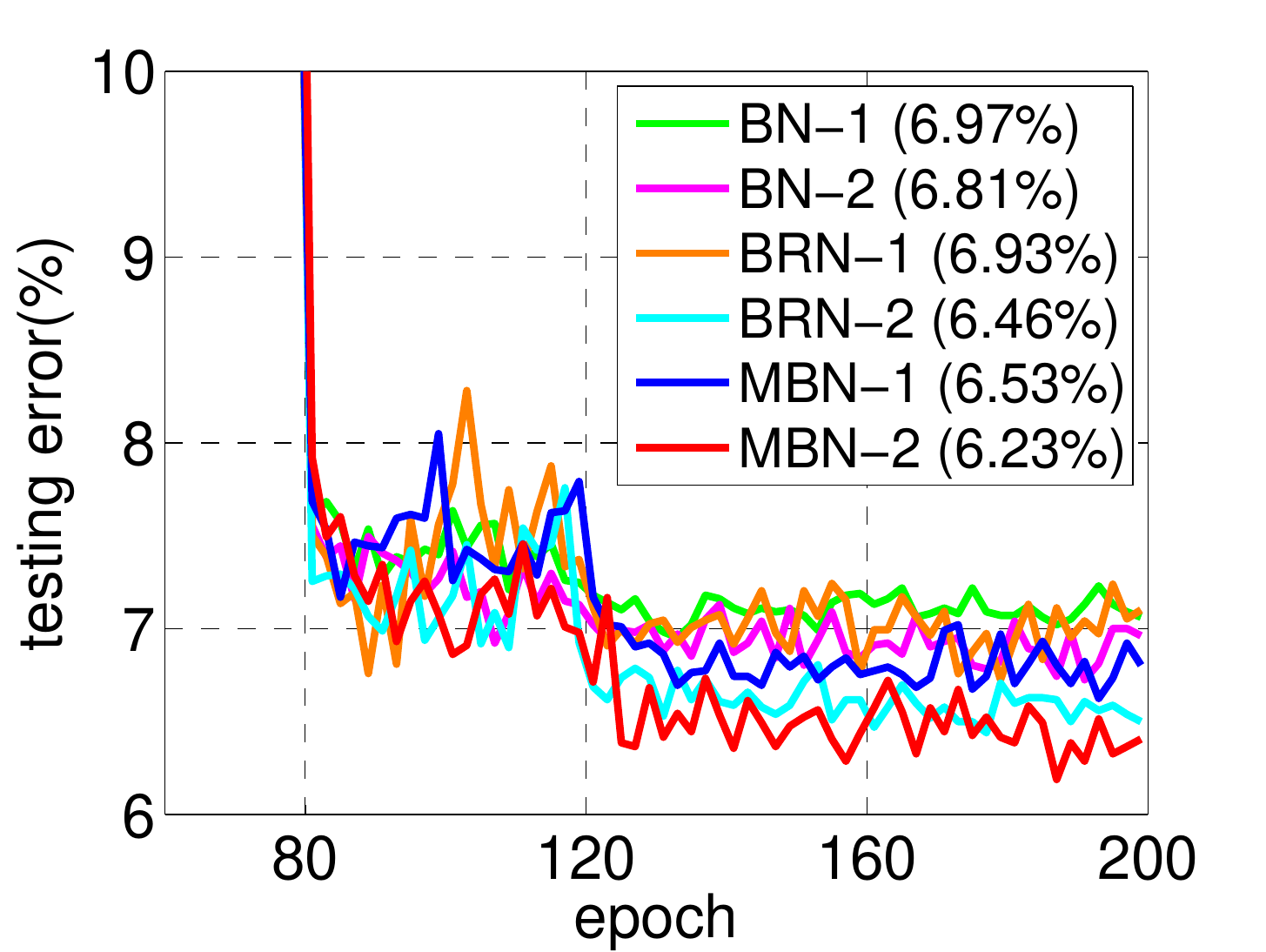}\label{fig:jointbp}
      }
      \subfigure[Training error.]{
      \includegraphics[width = 0.75\columnwidth]{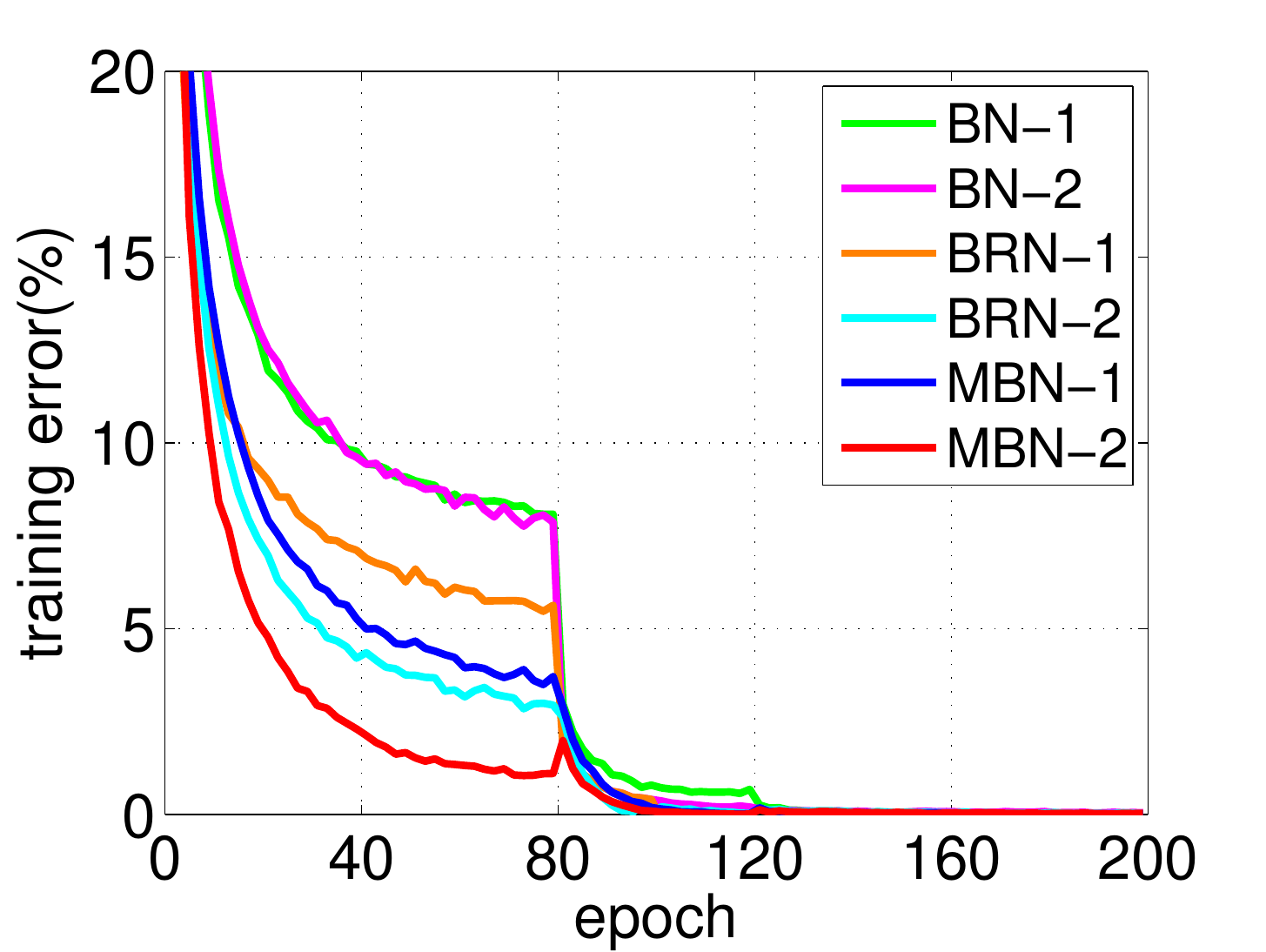}\label{fig:pairwisebp}
      }
  \caption{Demonstration of the effectiveness of Double-Forward Propagation for MBN. Based on ResNet-56 model, BN-1 and BN-2 denote the training with BN using single forward and double forwards, respectively; While MBN-1 and MBN-2, BRN-1 and BRN-2 have the same definition.}\label{fig:BP}
  \label{fig:doubleupdate}
\end{figure*}

\section{Experiments}
In this section, we evaluate the proposed MBN method in image classification tasks.
We apply MBN on several well-known models, including VGG~\cite{krizhevsky2012imagenet} and ResNet~\cite{he2016deep}.
Note that the original VGG models proposed in \cite{krizhevsky2012imagenet} do not have BN layer.
However, in many popular deep learning packages such as Tensorflow, Torch and PyTorch, they have a so called ``VGG-BN" model in their canonical model zoo, which add a BN layer after each convolutional layer in the original VGG-16 model. We adopt this VGG with BN model for comparisons.
For each model, we simply replace their original BN layers with the proposed MBN layers and keep the other parts unchanged. Several state-of-the-art methods are adopted as baselines, including BN~\cite{ioffe2015batch}, BRN\footnote{We implement BRN ourselves in PyTorch.}\cite{ioffe2017batch} and NormProp~\cite{arpit2016normalization}.

For fair comparisons, we follow the experimental settings in~\cite{he2016deep} for in all the experiments.
All compared models are implemented based on PyTorch.
Without special specification, we train the models through SGD with a mini-batch size of 128.
The momentum for SGD is 0.9 and the weight decay is set to ${10}^{-4}$.
The learning rate is initialized as 0.1 and then is divided by 10 at 40\% and 60\% of the training procedure, respectively.
For MBN methods, we first set the parameter $\lambda=0.1$ and then increase it to $0.5$ and $0.9$ at 40\% and 60\% of the training procedure, which is referred to as $\lambda=\{0.1,0.5,0.9\}$. And the decaying parameter $\eta$ in Eqn.(\ref{eq:decay}) is set to 0.9.
All the experiments are conducted on a GPU Server with one Titan X GPU.

\textbf{Datasets.} In the experiments, three benchmark datasets are used: CIFAR-10, CIFAR-100~\cite{krizhevsky2009learning} and ImageNet~\cite{russakovsky2015imagenet}.
CIFAR-10 contains 10 classes of 32x32 natural color images, each with 5,000 training samples and 1,000 testing samples.
For CIFAR-100, it has 100 classes, each of which has 500 training samples and 100 testing samples.
ImageNet contains 128 million high-resolution images belonging to 1000 categorise, and has become the canonical dataset for image classification, object detection and localization over the years.
The form of data augmentation consists of generating image translations and horizontal reflections~\cite{he2016deep}.
All the experiments are performed with 200 training epochs.

\subsection{Demonstration of Double-Forward Propagation}\label{sec:double-forward}
To verify the effectiveness of the proposed Double-Forward scheme, we apply it to train ResNet-56 for BN, BRN and MBN. Note that NormProp conduct normalization with model parameters, but the second forward propagation cannot change the model parameters. Therefore this scheme will have no effect on NormProp in the second forward propagation. The convergence results are shown in Figure~\ref{fig:doubleupdate}. We observe that first of all, MBN with single forward scheme (MBN-1) yields better performance than BN with whether single or Double-Forward scheme (6.53\% vs. 6.97\% and 6.81\%), and MBN with Double-Forward scheme (MBN-2) generates the best performance among all comparisons.

Furthermore, both MBN and BRN benefit greatly from the Double-Forward scheme at training and testing stage, but BN does not gain any improvement from this scheme during training
and merely improves negligibly at testing stage. For BN, it only takes the mean and variance of current batch during training. In this situation, forward once or twice makes no difference. However, moving mean and moving variance are used for BN during testing, which can be more accurate when the Double-Forward scheme is applied. In contrast, MBN and BRN gain much from this scheme in both training and testing, because they additionally utilizing the statistics of multiple previous batches to conduct the normalization, therefore the Double-Forward Propagation scheme can keep their current normalization statistics up-to-date.

For example, on ResNet model with 56 layers, MBN-2 achieves 0.30\% improvement in terms of testing error compared with MBN-1, and BRN-2 decreases the testing error by 0.47\% compared with BRN-1. Similar observations can be obtained on ResNet with 20 layers.
To be clear, in the following experiments, if not explicitly mentioned, MBN and BRN are set to use the Double-Forward Propagation scheme while BN and NormProp are set to use the single forward scheme. Note that the number of SGD updates for all the methods is kept the same for fair comparisons.

\begin{figure*}[t]
\centering

\subfigure[ResNet-20]{
\includegraphics[width = 0.66\columnwidth]{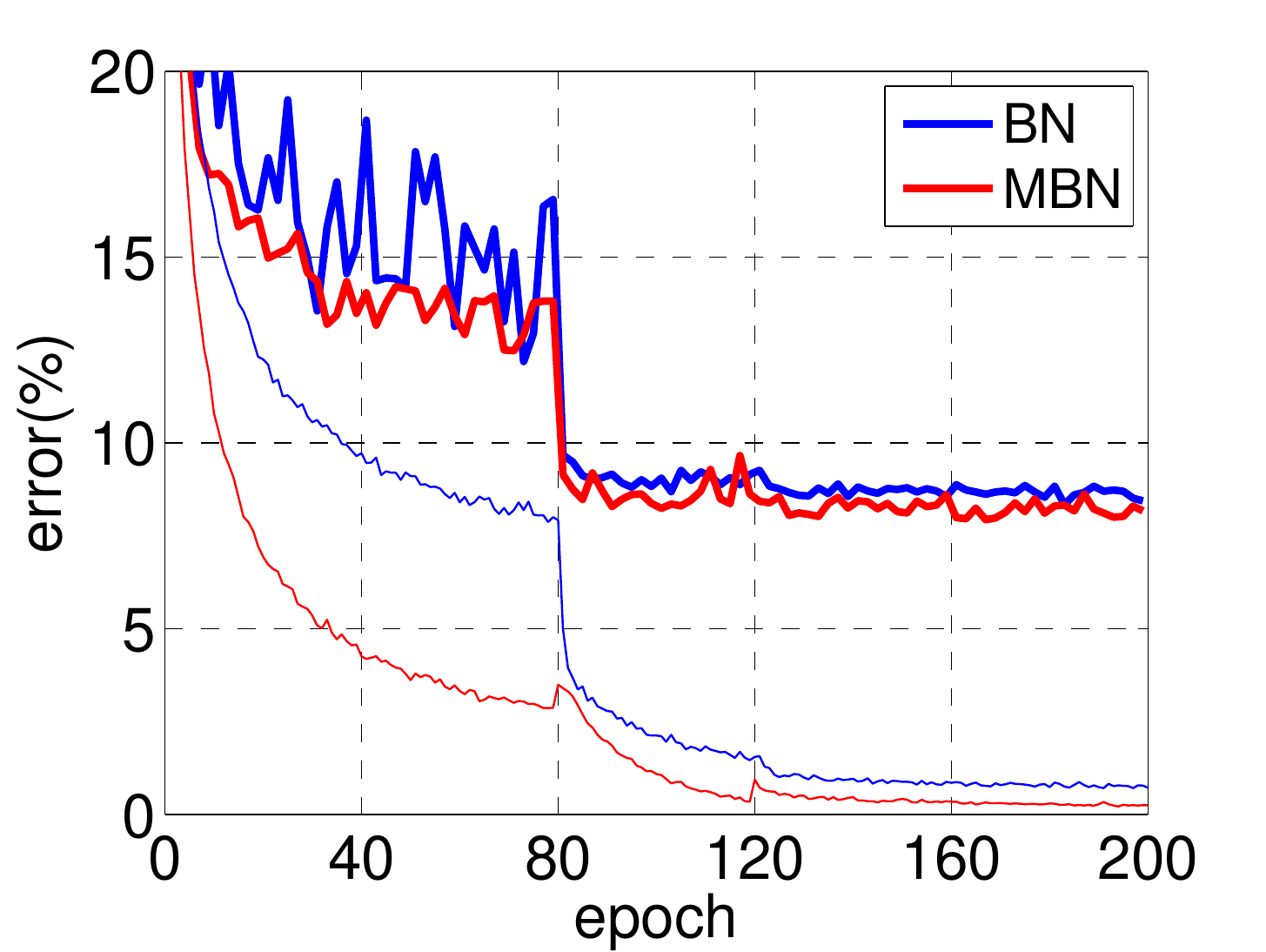}\label{fig:jointbp}
}
\subfigure[ResNet-56]{
\includegraphics[width = 0.66\columnwidth]{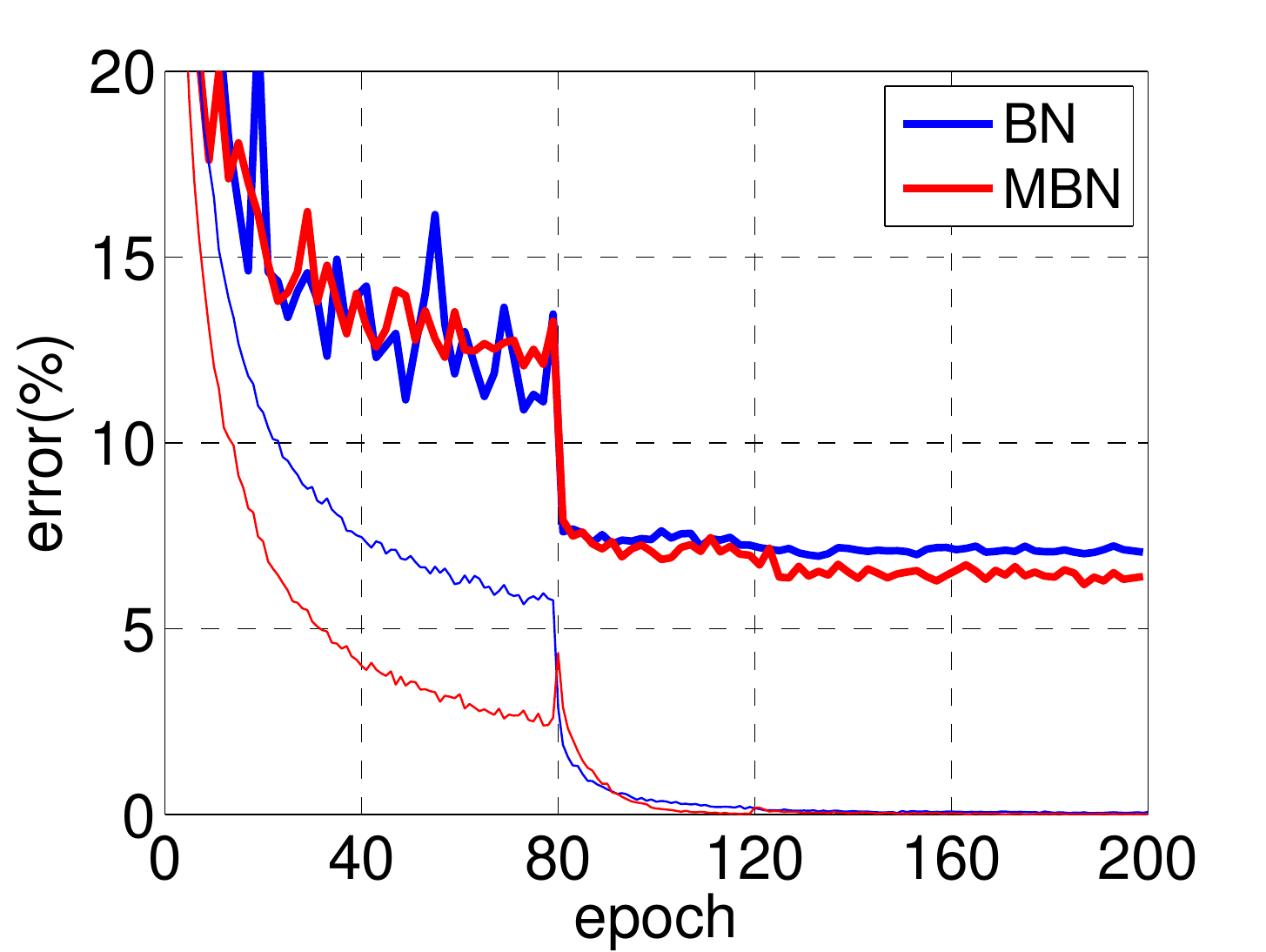}\label{fig:pairwisebp}
}
\subfigure[VGG-16]{
\includegraphics[width = 0.66\columnwidth]{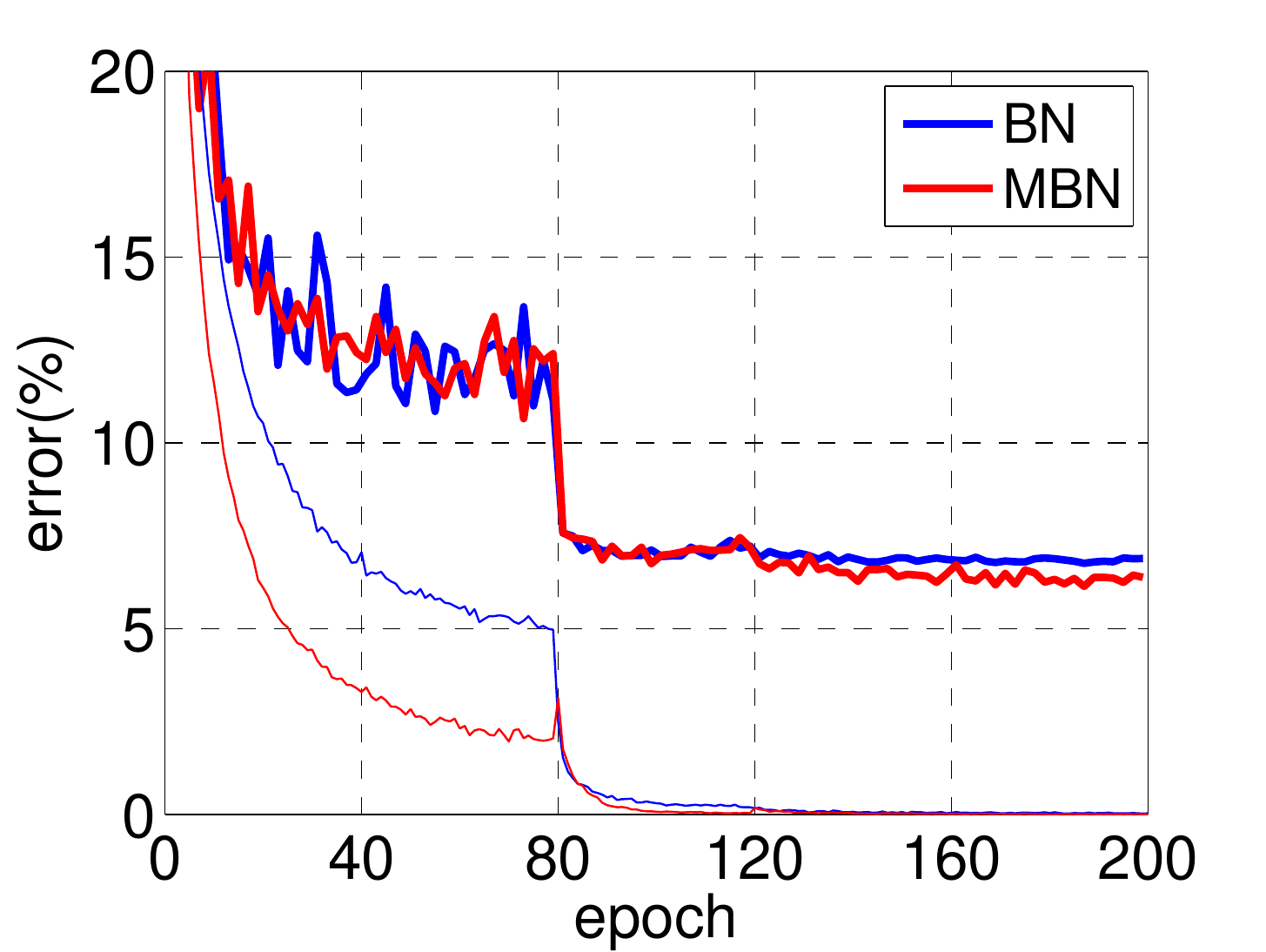}\label{fig:pairwisebp}
}

\caption{Comparisons of MBN and BN when applied on different deep models. Thin curves denote the training error while the bold curves denote the testing error.}\label{fig:BP}
\label{fig:cifar10}
\end{figure*}

\subsection{Performance on Image Classification}
\textbf{\textit{Comparisons on CIFAR-10}}. Firstly we study the convergence behavior and the evolution of testing errors of the three aforementioned models trained with MBN and BN, respectively. From Figure~\ref{fig:cifar10}, we have the following observations.
\textbf{First}, the models trained with MBN consistently outperform their BN based counterparts. Interestingly, the testing error of MBN based models decreases at the 120\textit{th} epoch where the learning rate is changed to $10^{-3}$, while the BN based models do not show such distinct improvements. This illustrates that MBN has unearthed the generalization potential of deep models.
\textbf{Second}, the abrupt increase of parameter $\lambda$ at 80\textit{th} epoch and 120\textit{th} epoch can raise the training error rates for a small moment, after that the error rates decrease rapidly, demonstrating that the MBN based models are capable to fit the updated hyper-parameters of memorized statistics in a very small number of iterations.
\textbf{Third}, the models trained with MBN generally have lower training error than those trained with BN.


\begin{table}[tbp]
  \centering
  \caption{Performance comparisons of MBN and different normalization methods on CIFAR-10, where `-' stands for the absence of results.}
    \begin{tabular}{c|c|c|c|c}
    \hline
    method & NIN & ResNet-20 & ResNet-56 & VGG \\
    \hline
    NormProp   &   7.25    &   -   &  - &   11.18  \\
    BN   &   7.54    &   8.41    &   6.97    &   6.75  \\
    BRN   &   -    &   8.18    &   6.46    &   6.64  \\
    MBN   &   -    &    7.93   &   \textbf{6.23}    &   6.43   \\
    \hline
    \end{tabular}%
  \label{tab:cifar10}%
\end{table}%

We further compare MBN with other baselines. The model details and testing errors are recorded in Table~\ref{tab:cifar10}. From the table, we can see that ResNet-56 trained with the proposed MBN yields the best performance. Specifically, MBN reduces the testing error by $0.48\%$ and $0.74\%$ for ResNet-20 and ResNet-56, respectively. On VGG model, the MBN method still outperforms every other comparison.
For BRN method, the moving averages in BRN can be very close to the batch statistics $\mu_{\mathcal B}, \sigma_{\mathcal B}$ because of the ``fairly high" decay rate $\theta$ in Eqn.~({\ref{moving_aveage}}), which makes the result of BRN quite close to BN.
Nevertheless, for NormProp method, it cannot be directly applied on residual module and it is only suitable for shallow models like NIN~\cite{arpit2016normalization}. Equipped with NormProp, even a 16 layers model, i.e. VGG, would very likely to explode and suffers a large reduction in performance.
These results demonstrate the superior performance of the proposed MBN method.

\textbf{\textit{Comparisons on CIFAR-100}}.
We follow the same experimental settings in CIFAR-10 experiments for all the compared methods. The results are recorded in Table~\ref{tab:cifar100}.
Among all the evaluated models, MBN based models yield the best performance. Compared with BN, MBN exhibits $0.5\%$ and $0.3\%$ performance improvements on ResNet-20 and ResNet-56, respectively. For the VGG model, when trained with MBN method, it achieves a reduction of $1\%$ in testing error, which can be a significant improvement. Note that BRN method is also able to boost the performance compared to BN, but the improvements it brings are not as much as MBN method does.

\begin{table}[tbp]
  \centering
  \caption{Performance comparisons of MBN and different normalization methods on CIFAR-100, where `-' stands for the absence of results.\\}
    \begin{tabular}{c|c|c|c|c}
    \hline
    method & NIN & ResNet-20 & ResNet-56 & VGG \\
    \hline
    NormProp    &   29.24    &   -    &  -  &   39.55  \\
    BN   &  30.26   &   32.28    &   29.27    &   27.80  \\
    BRN  &  -   &   32.04    &   29.17    &   27.44  \\
    MBN   &  -  &    31.75   &   28.97    &   \textbf{26.79}   \\
    \hline
    \end{tabular}%
  \label{tab:cifar100}%
\end{table}%

\begin{figure*}[t]
  \centering
      \subfigure[Testing error.]{
      \includegraphics[width = 0.77\columnwidth]{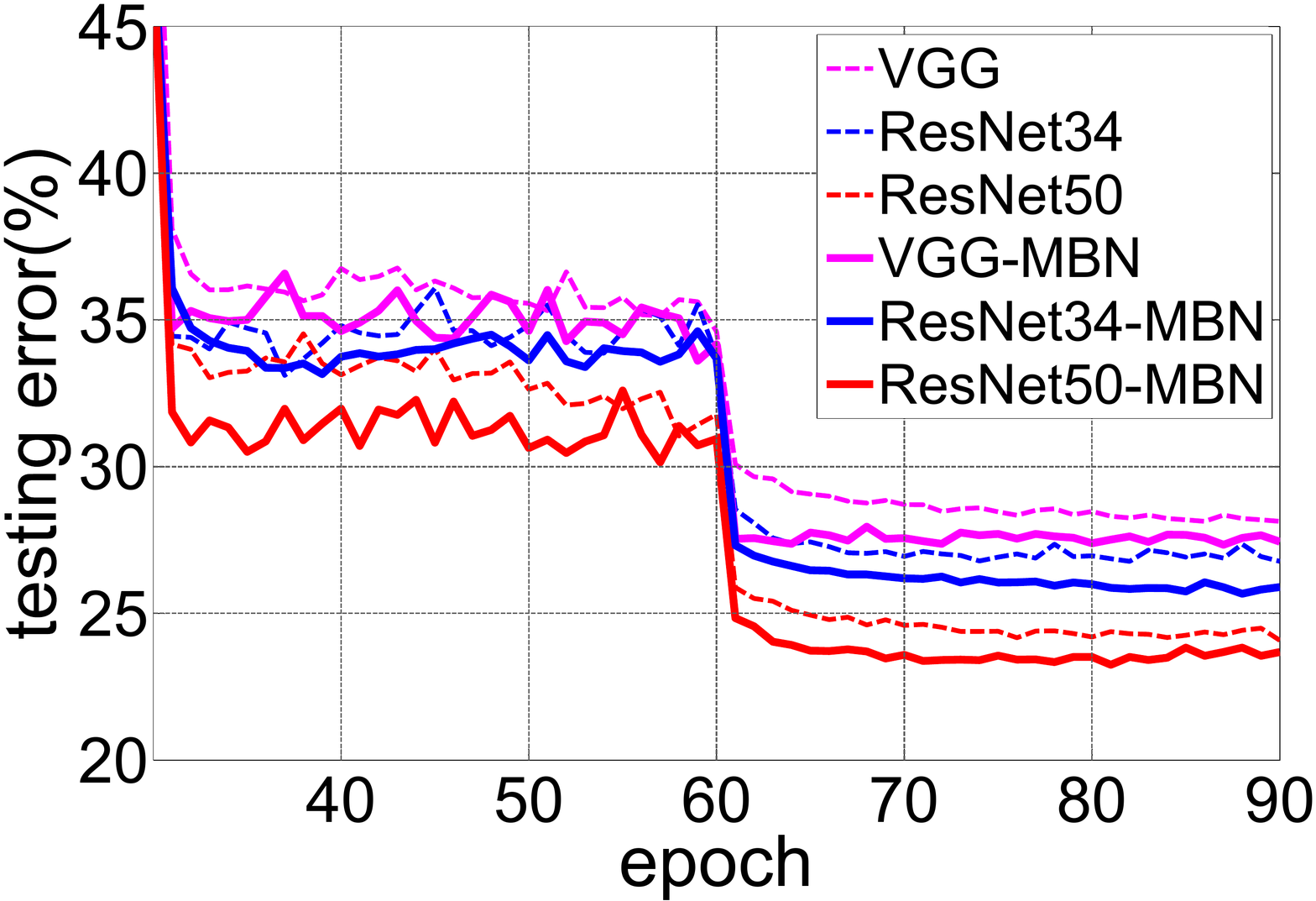}\label{fig:jointbp}
      }
      \subfigure[Training error.]{
      \includegraphics[width = 0.77\columnwidth]{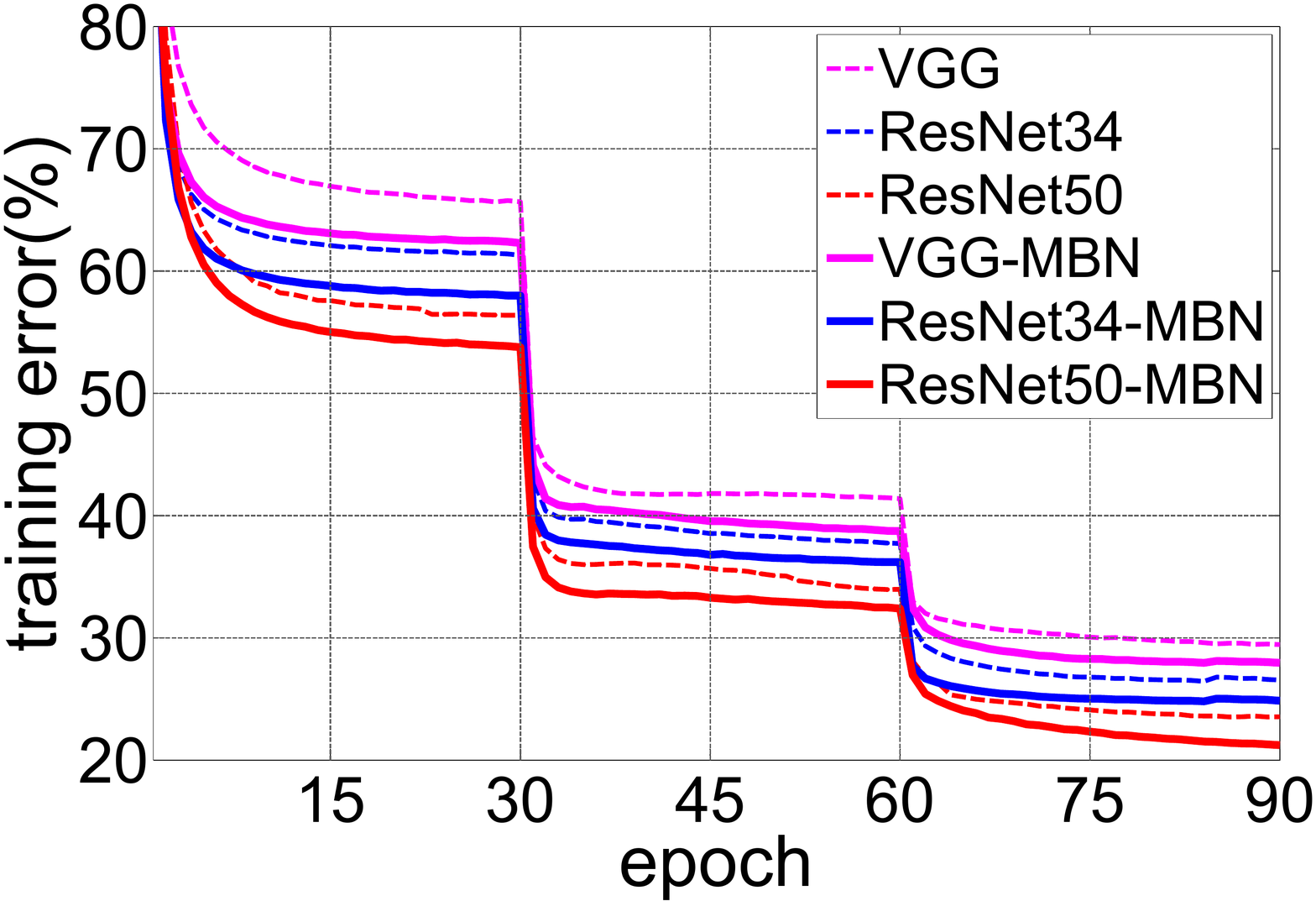}\label{fig:pairwisebp}
      }
  \caption{Results on ImageNet. Solid curves denote the MBN-based methods while dash curves denote the BN-based methods }\label{fig:BP}
  \label{fig:imagenet}
\end{figure*}

\textbf{\textit{Comparisons on ImageNet}}.
We evaluate the proposed method on a more challenging ILSVRC2015 dataset. We adopt two state-of-the-art models GoogLeNet~\cite{szegedy2015going} and PReLU-net~\cite{he2015delving}. To evaluate the proposed MBN method, We train VGG and ResNet of different depth using BN and MBN. The results are shown in table~\ref{tab:imagenet}. Several models are taken into consideration, including ResNet-18, ResNet-34, ResNet-50 and VGG with 16 layers. As shown in Table~\ref{tab:imagenet}, MBN outperforms BN on every compared model, and shows significant improvements. More specifically, MBN decreases the top-1 testing error rate by 0.74\% and 0.88\% on VGG and ResNet-34, respectively. These results verify the effectiveness of the proposed method.

We also plot the convergence curves of models trained with BN and MBN in Figure~\ref{tab:imagenet}.
Different from CIFAR datasets, we do not observe the abrupt increase in training curves when increasing the hyper-parameter $\lambda$ on ImageNet. A reasonable explanation is that the enormous training samples in ImageNet can make the memorized statistics more robust to the hyper-parameter $\lambda$.
\begin{table}[htbp]
  \centering
  \caption{Performance comparisons on ImageNet (single-crop testing).}
    \begin{tabular}{c|c|c|c|c}
    \hline
    \multirow{2}[0]{*}{network} & \multicolumn{2}{c|}{top-1 error (\%)} & \multicolumn{2}{c}{top-5 error (\%)} \\
    \cline{2-5}
        &  BN  & MBN  &  BN  & MBN \\
    \hline
    GoogLeNet &   27.31   & -     & 9.15  & -\\
    PReLU-net & 24.27 &  -  & 7.38 & - \\
    VGG &   28.07   &  27.33     & 9.33  & 8.91\\
    ResNet-18 & 30.43     & 29.79     & 10.76  & 10.23 \\
    ResNet-34 & 26.73    & 25.85     & 8.74  & 8.37\\
    ResNet-50 & 24.01    & 23.67     & 7.02   & 6.74\\
    \hline
    \end{tabular}
  \label{tab:imagenet}
\end{table}

\subsection{Performance Comparison with Small Mini-batches}
We investigate the effectiveness of MBN with small mini-batches by setting the mini-batch size to the values in $\{8, 16,32, 64, 128\}$ to train deep models on CIFAR-10. For fair comparisons, we keep the same setting to previous experiments but constrain the training stages to have the same number of SGD iterations.
As shown in Table~\ref{tab:smallbatch}, the models trained with MBN show significant improvements over their BN counterparts with all kinds of batch size, which illustrates the effectiveness of MBN with small batch sizes.

\begin{table}[htbp]
  \centering
  \caption{Testing errors of BN and MBN with different batch sizes.\\}
    \begin{tabular}{c|c|c|c}
    \hline
    \multirow{2}[0]{*}{network} & \multirow{2}[0]{*}{batch size} & \multicolumn{2}{c}{testing error (\%)} \\
    \cline{3-4}
        &       & BN  & MBN \\
    \hline
    \multirow{5}[0]{*}{ResNet-56} & 8     & 19.64     & 15.37 \\
                                  & 16     & 13.26     & 11.87 \\
                                  & 32     & 10.45     & 9.37 \\
                                  & 64     & 8.60     & 7.79 \\
                                  & 128     & 6.97     & 6.23 \\
    \hline
    \end{tabular}
  \label{tab:smallbatch}
\end{table}

\section{More Discussions}
\subsection{Effect of the Parameter $\lambda$ in Eqn. (\ref{eq:decay})}
The choice of the weight decaying parameter $\lambda$ is critical to the performance of MBN based models. We here study the influence of $\lambda$ on the performance of MBN. We can either set $\lambda$ to a fixed value or adjust it adaptively w.r.t. training epochs. Different settings and the relevant performance are recorded in Table~\ref{tab:lambda}. We observe that choosing a fixed value, (e.g. $\lambda=0.1$, $0.5$ or $1$) fails to produce good results. Actually, since the distribution shift is different in different stages, a fixed value of $\lambda$ is not a reasonable choice.
Here, we suggest setting a small value at the beginning and then increasing it along training time. For example, in our comparison experiments above,  we set $\lambda=0.1$ at the beginning stage, 0.5 and 0.9 at 40\% and 60\% of the training procedure, respectively. With this adaptive setting of $\lambda$, MBN  performs much better performances compared to other baselines.

\begin{table}[htbp]
  \centering
  \caption{The effect of the parameter $\lambda$ on MBN.\\}
    \begin{tabular}{c|c|c}
    \hline
    network & settings of $\lambda$ & testing error (\%) \\
    \hline
    \multirow{5}[0]{*}{ResNet-56}
          & $\lambda=\{0\} $    & 6.97 \\
          & $\lambda=\{0.1\} $    & 6.87 \\
          & $\lambda=\{0.5\} $   & 6.45 \\
          & $\lambda=\{0.9\} $    &  6.81 \\
          & $\lambda=\{0.1, 0.5, 0.9\}$    &  \textbf{6.23} \\
    \hline
    \end{tabular}%
  \label{tab:lambda}%
\end{table}%

\subsection{Influence of Memory Size}
In this experiment, we investigate the influence of the memory size $k$ on the performance of MBN. We conduct this evaluation using ResNet with 20 layers on CIFAR-10. The results are represented in Table~\ref{tab:memorysize}. From the table, we see that the performance of MBN is relatively stable across different memory sizes. In other words, carefully adjusting the memory size is not necessary.

\begin{table}[htbp]
  \centering
  \caption{The influence of memory size on MBN.\\}
    \begin{tabular}{c|c|c}
    \hline
    network & memory size & testing error (\%) \\
    \hline
    \multirow{3}[0]{*}{ResNet-20}
          & 10    & 7.87  \\
          & 20    &  7.80 \\
          & 40    &  7.83 \\
    \hline
    \end{tabular}%
  \label{tab:memorysize}%
\end{table}%

\section{Conclusion}
In this paper, we propose a Memorized Batch Normalization (MBN) method for training DNNs. In contrast to BN which relies on a single batch to perform data normalization, MBN considers multiple recent batches to obtain more accurate estimations of the statistics. We also propose several simple techniques to alleviate the \emph{Distribution Shift} among batches. Unlike BN, the proposed MBN method exhibits consistent behaviors in both training and inference, and it can play an effective role in dealing with small mini-batch size cases.
The extensive experiments have demonstrated the effectiveness of the proposed methods.

\section{Acknowledgements}
This work was supported by National Natural Science Foundation of China (Grand No. 61502177 and 61602185), Recruitment Program for Young Professionals,
Guangdong Provincial Scientific and Technological Funds (Grand No. 2017B090901008 and 2017A010101011), Fundamental Research Funds for the Central Universities (Grand No. D2172500 and D2172480), Special Planning Project of Guangdong Province (Grand No. 609055894069) and CCF-Tencent Open Research Fund.

{
\bibliographystyle{aaai}
\bibliography{bn}
}

\end{document}